\documentclass[letterpaper, 10 pt, conference]{ieeeconf}  

\IEEEoverridecommandlockouts                              

\usepackage{cite}
\usepackage{graphics} 
\usepackage{subfigure}
\usepackage{epsfig}
\usepackage{booktabs}
\usepackage{makecell}
\usepackage{multirow}
\usepackage{times} 
\usepackage{amsmath} 
\usepackage{amssymb}  

\title{\LARGE \bf
An MPC-based Optimal Motion Control Framework for Pendulum-driven Spherical Robots
}

\author{Tao Hu$^{1}$, Xiaoqing Guan$^{1}$, Yixu Wang$^{1}$, Yifan Liu$^{1}$, Bixuan Zhang$^{1}$, Boyu Lin$^{1}$, You Wang$^{1,2}$ and Guang Li$^{1}$
\thanks{This work is supported by the Fundamental Research Funds for the Central Universities 226-2022-00086.}%
\thanks{$^{1}$Authors are with State Key Laboratory of Industrial Control Technology, Institute of Cyber Systems and Control, Zhejiang University, Hangzhou, China (email: hutao@zju.edu.cn, xiaoqing\_guan@zju.edu.cn, yixuwang@zju.edu.cn, yifanliu@zju.edu.cn, Bixuan\_zhang@zju.edu.cn, Boyu\_Lin@zju.edu.cn, king\_wy@zju.edu.cn, guangli@zju.edu.cn)}%
\thanks{$^{2}$Corresponding author: You Wang}%
}

\begin{document}

\maketitle
\thispagestyle{empty}
\pagestyle{empty}

\begin{abstract}

Motion control is essential for all autonomous mobile robots, and even more so for spherical robots. Due to the uniqueness of the spherical robot, its motion control must not only ensure accurate tracking of the target commands, but also minimize fluctuations in the robot's attitude and motors' current while tracking. In this paper, model predictive control (MPC) is applied to the control of spherical robots and an MPC-based motion control framework is designed. There are two controllers in the framework, an optimal velocity controller ESO-MPC which combines extend states observers (ESO) and MPC, and an optimal orientation controller that uses multilayer perceptron (MLP) to generate accurate trajectories and MPC with changing weights to achieve optimal control. Finally, the performance of individual controllers and the whole control framework are verified by physical experiments. The experimental results show that the MPC-based motion control framework proposed in this work is much better than PID in terms of rapidity and accuracy, and has great advantages over sliding mode controller (SMC) for overshoot, attitude stability, current stability and energy consumption.

\end{abstract}

\keywords
Model predictive control, motion control, spherical robots
\endkeywords

\section{INTRODUCTION}

Spherical robots are a novel type of mobile robots with better sealability, environmental adaptability and energy efficiency ratio than wheeled and legged robots \cite{ballreview1}. Furthermore, the overall spherical shape prevents them from flipping over. Therefore, spherical robots have a wide range of applications in security, exploration, search and rescue, and other fields. However, since the spherical robot is a nonlinear, under-actuated, and nonholonomic system, the complexity of motion control severely limits its development. Because the control strategies of spherical robots with different structures vary \cite{ballreview2}, this paper focuses on the motion control of pendulum-driven spherical robots.

For autonomous mobile robots, perception, localization, cognition, navigation and locomotion are all essential\cite{mobilerobotsreview1}. Among them, the locomotion module is responsible for solving motion control problems, and control the robot to follow the target values given by the navigation module. For typical mobile robots such as wheeled robots, the robot itself is a stable platform and motion has less impact on perception, so their motion control modules are more concerned with reacting quickly, precisely and stably. For spherical robots, however, things are much more complex. Firstly, spherical robots move by rolling, and the sensors are solidly attached to the sphere, so the motion of robots directly affects the quality of the sensor data. The rapid shaking of sensors (especially in-plane rotation) can lead to problems of motion blur \cite{motionblur1} (also known as motion distortion in LiDAR \cite{motiondistortion1}). Secondly, the spherical robot is a ball-pendulum system \cite{backstep1}, and the motors control the pendulum to realize movement. Frequent fluctuations in the motors' current will cause high-frequency vibration of the pendulum, which will affect the structure's stability and robot's lifespan. As a result, the motion control of spherical robots must achieve fast and accurate tracking of the target while ensuring the stability of the robot's attitude and motors' current as much as possible.

\begin{figure}[h]
\centering
\subfigure[physical view of the robot]{
\includegraphics[width=0.42\linewidth]{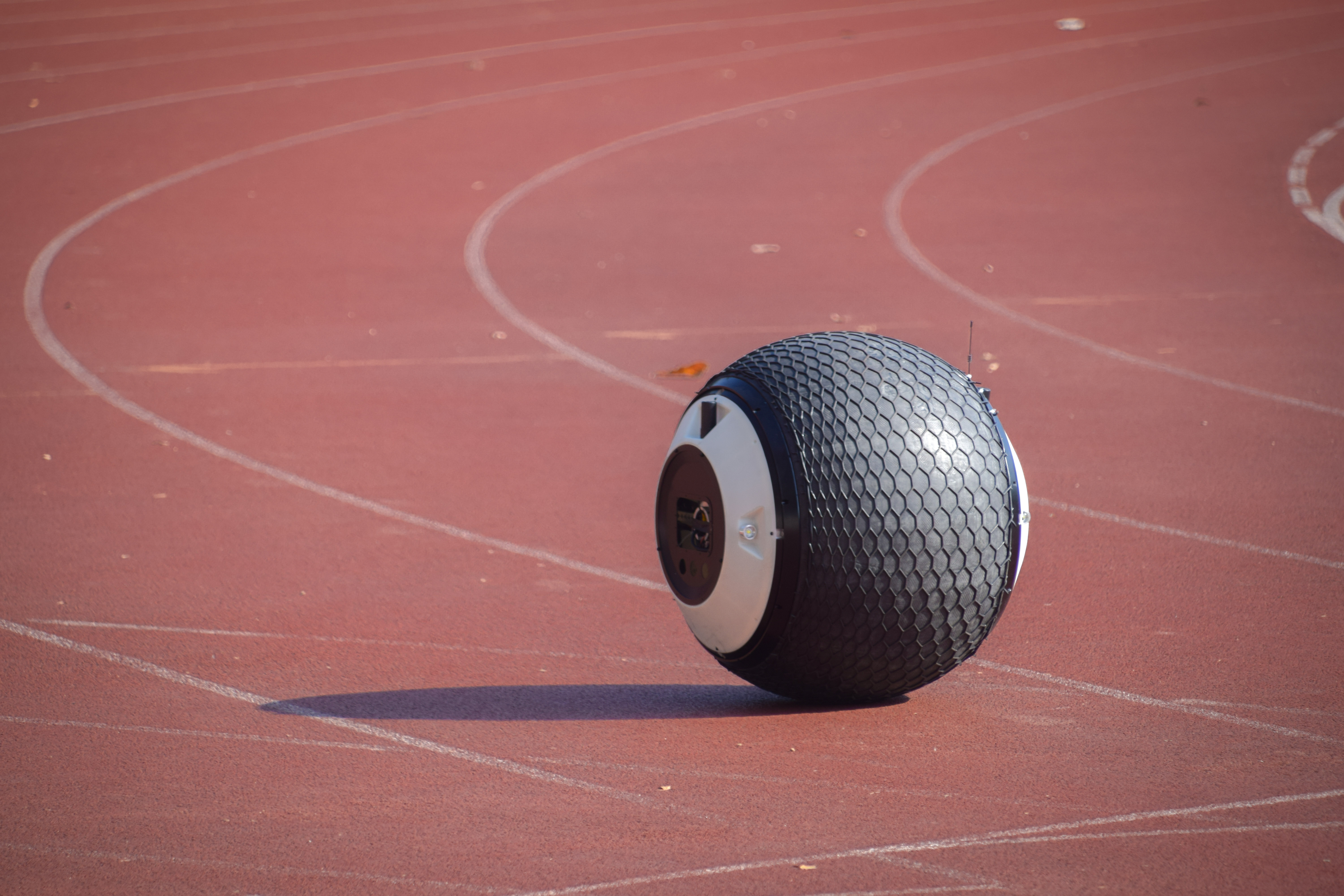}
}
\quad
\subfigure[simplified model of the robot]{
\includegraphics[width=0.42\linewidth]{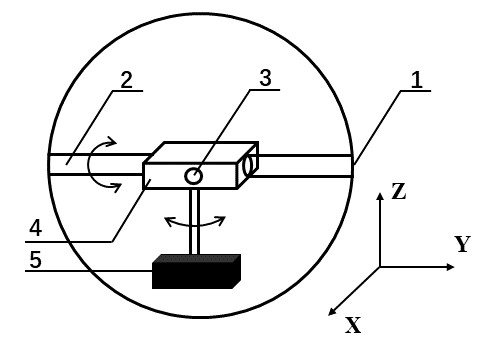}
}
\caption{The spherical robot used in this study. In (b), Component 1 is the spherical shell, 2 is the longitudinal axis, 3 is the transversal axis, 4 is the frame and 5 is the pendulum.}
\label{Fig-robot}
\vspace{-0.5cm}
\end{figure}

Although spherical robots' motion control is challenging, researchers have achieved some progress. Scholars have designed several model-based controllers for spherical robots, such as state feedback controller \cite{statefeedback}, fuzzy controller \cite{fuzzy1, fuzzy2}, and model reference adaptive controller \cite{modelreference}. Backstepping techniques have also been used in some studies. In \cite{backstep1}, backstepping was used to control the robot's motion, including position and orientation. Y. Cai combined a hierarchical sliding mode controller with a fuzzy guidance scheme through a backstepping strategy to achieve path tracking \cite{backstep2}. Additionally, a two-state trajectory tracking controller was conducted by integrating a shunting model of neurodynamics and Lyapunov’s direct method \cite{neurodynamic}. However, the above controllers have only been validated in simulation and we are not sure if they will work properly in real systems and environments. In our previous work, we built a pendulum-driven spherical robot, as in Fig. \ref{Fig-robot}, developed and tested algorithms on it. Fuzzy-PID \cite{fuzzy} as well as SMC \cite{hsmc1,hsmc2,hsmc3} were applied to solve the motion control problem. Unfortunately, there are still some problems with these controllers: Fuzzy-PID responds slowly, and it cannot control the robot to accurately track the target when the target is constantly changing. As for SMC, although its stability, rapidity, and accuracy have been verified, it cannot consider the robot's attitude well, causing drastic fluctuations in robot's attitude when tracking, which has a serious impact on perception, and there are frequent and abrupt changes in the motors' current due to the algorithm's chattering problem \cite{chattering}.

In order to solve the above problems, this paper views the motion control of spherical robots as a multi-objective optimal control problem and introduces model predictive control for its solution. Model predictive control (MPC) is an optimal control method with the advantage of systematically handling multi-variable system dynamics, constraints and conflicting control objectives \cite{mpcprinciple1, mpcprinciple2}. Despite the fact that MPC has not yet been used in the control of pendulum-driven spherical robots, the algorithm has been successfully applied to a variety of other robots \cite{fixwingUAV, Quadrotor, underwater, quadrupedrobot, wheeledlegged, vehicles}. In fixed-wing unmanned aerial vehicles (UAV), linear model predictive control (LMPC) combined with feedback linearization was used as a low-level controller \cite{fixwingUAV}. A time-varying LMPC controller was applied in quadruped robots to get the desired ground reaction force \cite{quadrupedrobot}. As in autonomous ground vehicles (AGV), an MPC-based obstacle avoidance framework was proposed, including a high-level trajectory generator and a low-level MPC controller \cite{vehicles}. These instances serve as a guidance for using MPC on spherical robots.

Nevertheless, the complexity of the whole system limits the direct use of MPC. In this paper, the whole system is decomposed into two subsystems, and then MPC is improved according to the characteristics of the subsystems respectively. Finally, a velocity controller (ESO-MPC) is designed by combining an extend states observer and MPC, and an MPC controller whose weights vary with phase (PWMPC) based on the dynamics is used as an orientation controller. The two controllers successfully establish an optimal motion control framework for spherical robots. The main contributions of this paper are as follows:
\begin{itemize}
  \item [$\bullet$]
  Based on the spherical robot's characteristics, the effects of robot's motion on perception and motors' current on mechanical structure are taken into account in the motion control of spherical robots for the first time.
  \item [$\bullet$]
  MPC is first applied to the motion control of pendulum-driven spherical robots. The complicated robot system is decomposed into two subsystems to apply MPC separately, which achieves optimal control while improving computing efficiency.
  \item [$\bullet$]
  The performance of the proposed MPC-based framework is verified on a real robot. The experiments show that, whether it is used for velocity control, orientation control, or both, the MPC-based controllers perform better than PID and SMC controllers.
\end{itemize}

This paper is organized as follows : The spherical robot's whole-body dynamic model is established in Section \uppercase\expandafter{\romannumeral2}. Section \uppercase\expandafter{\romannumeral3} develops an MPC-based optimal motion control framework for spherical robots. In section \uppercase\expandafter{\romannumeral4}, physical experiments are carried out to compare the performance of each controller. The summary and outlook are in section \uppercase\expandafter{\romannumeral5}.

\section{THE DYNAMIC MODEL}

The robot studied in this work is shown in Fig. \ref{Fig-robot}, and is primarily made up of a spherical shell, an inner frame and a 2-DOF pendulum. It moves when the pendulum swings around the x and y axes under the control of the motors. The sensors required are mounted on the flanges on both sides of the robot, which are fixed to the longitudinal axis. 

In order to introduce MPC in the spherical robot's motion control, we must first model the robot's dynamics. The variables used are presented in Table \uppercase\expandafter{\romannumeral1}.
\begin{table}[h]
\renewcommand{\arraystretch}{1.1}
\setlength{\tabcolsep}{3pt}
\vspace{-0.15cm}
\setlength{\abovecaptionskip}{-0.3cm}
\caption{Nomenclature}
\label{Nomenclature}
\begin{center}
\begin{tabular}{cc}
\toprule
Symbol & Description\\
\midrule
$m_s, m_f, m_p$ & Mass of the shell, frame and pendulum\\
$I_{s}, I_{f}, I_{p}$ & Moment of inertia of the shell, frame and pendulum \\
$\alpha, \beta$ & Swing angle of the pendulum about y and x axes\\
$\theta, \phi$ & Swing angle of the shell about y and x axes\\
$F_{fx}, F_{fy}$ & Friction between the shell and ground along x and y axes\\
$\tau_1, \tau_2$ & Torque of the long-axis motor and short-axis motor\\
$x$ & Distance that the spherical robot moves along the x-axis\\
$r$ & Radius of the spherical robot\\
$l$ & Distance between the frame and the pendulum\\
$\zeta$ & Viscous damping coefficient\\
\bottomrule
\end{tabular}
\end{center}
\vspace{-0.5cm}
\end{table}

By using the Euler-Lagrange method, the dynamic model of the robot is obtained, as in \eqref{euler-lagrange}, where $L$ denotes the Lagrange function and $\Psi$ is the Rayleigh's dissipation function. 
\begin{equation}
    \label{euler-lagrange}
    \frac{d}{d t}\left(\frac{\partial L}{\partial \dot{q}_{i}}\right)-\frac{\partial L}{\partial q_{i}}+\frac{\partial \Psi}{\partial \dot{q}_{i}}=\tau_{q_{i}}
\end{equation}

\noindent in which $\setlength{\arraycolsep}{2.0pt} q_i \in \boldsymbol{q} =  \left[\begin{array}{llll}
\alpha & x & \beta & \phi \end{array}\right]^{T}$, and $\tau_{q_{i}}$ is the related external torque.

After simplification, we can get the robot's whole-body dynamic model in the following form:
\begin{equation}
    \label{whole-body dynamics}
    \boldsymbol{M}(\boldsymbol{q})\ddot{\boldsymbol{q}}+\boldsymbol{N}(\boldsymbol{q}, \dot{\boldsymbol{q}}) = \boldsymbol{E} \boldsymbol{\tau}
\end{equation}

\noindent where
$$
\setlength{\arraycolsep}{2pt}
\boldsymbol{M}(\boldsymbol{q})\!=\!\left[\begin{array}{cccc}
I_{fy}\!+\!I_{py} & m_p l \cos \alpha & 0 & 0 \\
m_p r l \cos \alpha & M r\!+\!I_{sy} / r & 0 & 0 \\
0 & 0 & I_{px} & m_p r l \cos \beta \\
0 & 0 & m_p r l \cos \beta  & M r^{2}\!+\!I_{s x}\!+\!I_{f x}
\end{array}\right]
$$
$$
\setlength{\arraycolsep}{2pt}
\boldsymbol{N}(\boldsymbol{q}, \dot{\boldsymbol{q}})=\left[\begin{array}{c}
m_p g l \sin \alpha \cos \beta+\zeta(\dot{\alpha}+\dot{x} \cos \alpha / r) \\
-m_p r l \dot{\alpha}^{2} \sin \alpha+\zeta(\dot{\alpha} \cos \alpha+\dot{x} / r)+F_{f x} r \\
m_p g l \cos \alpha \sin \beta+\zeta\left(\dot{\beta}+\dot{\phi} \cos \beta\right) \\
-m_p r l \dot{\beta}^{2} \sin \beta+\zeta\left(\dot{\phi}+\dot{\beta} \cos \beta\right)+F_{f y} r
\end{array}\right]
$$
$$
\boldsymbol{E}=\left[\begin{array}{cccc}
1 & 1 & 0 & 0 \\
0 & 0 & 1 & 1
\end{array}\right]^{T}, \boldsymbol{\tau}=\left[\begin{array}{l}
\tau_{1} \\
\tau_{2}
\end{array}\right]
$$

\section{THE MPC-BASED CONTROL FRAMEWORK}

\begin{figure}[h]
\vspace{-0.3cm}
\centering
\setlength{\abovecaptionskip}{-0.1cm}
\includegraphics[width=1.0\linewidth]{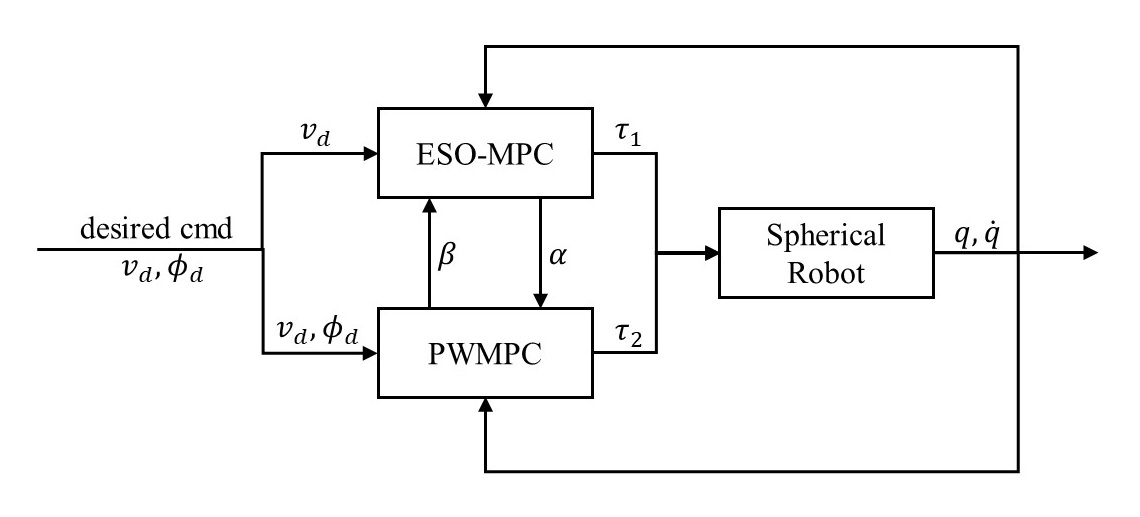}
\caption{The MPC-based motion control framework for spherical robots.}
\label{Fig-frame}
\vspace{-0.3cm}
\end{figure}

In this section, we first select linear model predictive control (LMPC) from the perspective of computational complexity (discussed in subsection A), then we design controllers for spherical robots using LMPC, including an optimal velocity controller ESO-MPC (shown in subsection B) and an optimal orientation controller PWMPC (seen in subsection C, D and E). Together, the two controllers form the optimal motion control framework for spherical robots, as shown in Fig. \ref{Fig-frame}.

\subsection{Linear Discrete Time Dynamics}

After obtaining the robot's dynamic model, an intuitive way to apply MPC is to directly use the discretized model as a predictive model and design a whole-body nonlinear model predictive controller (NMPC). However, NMPC is often computational intensive and time-consuming, which is the main problem restricting its application \cite{NMPC}. The average solution time of the whole-body NMPC is tested to be 0.06s, which does not meet the control frequency requirement of 50 Hz. (Tests are performed using the automatic differentiation package Cppad \cite{cppad}, and the arising nonlinear programming problems are solved with IPOPT \cite{ipopt}.)

In the case that the whole-body NMPC does not satisfy the requirements, we refer to previous work \cite{fuzzy1, hsmc3} and decompose the whole-body model into two sub-models as shown in equations \eqref{longitudinal-axis} and \eqref{transverse-axis}, where \eqref{longitudinal-axis} for the longitudinal-axis model and \eqref{transverse-axis} for the transverse-axis model. However, even the NMPC for the sub-models still falls short of the required solution time. As a result, we chose LMPC algorithms.
\begin{equation}
    \label{longitudinal-axis}
    \setlength{\arraycolsep}{2pt}
    \boldsymbol{M_p}(\boldsymbol{q_p}) \ddot{\boldsymbol{q_p}}+\boldsymbol{N_p}(\boldsymbol{q},\dot{\boldsymbol{q}}) = \boldsymbol{E_1} \tau_1
\end{equation}
\begin{equation}
    \label{transverse-axis}
    \setlength{\arraycolsep}{2pt}
    \boldsymbol{M_r}(\boldsymbol{q_r}) \ddot{\boldsymbol{q_r}}+\boldsymbol{N_r}(\boldsymbol{q},\dot{\boldsymbol{q}}) = \boldsymbol{E_2} \tau_2
\end{equation}

\noindent in which $\setlength{\arraycolsep}{2pt} \boldsymbol{q_p} \!=\! \left[\begin{array}{cc}
\alpha & x \end{array}\right]^{T}$, $\setlength{\arraycolsep}{2pt} \boldsymbol{q_r} \!=\! \left[\begin{array}{cc}
\beta & \phi \end{array}\right]^{T}$, $\setlength{\arraycolsep}{2pt} \boldsymbol{E_1} \!=\! \boldsymbol{E_2} \!=\! \left[\begin{array}{cc}
   1  & 1 \end{array} \right]^T$
$$
\setlength{\arraycolsep}{2pt} \boldsymbol{M}(\boldsymbol{q}) = \left[\begin{array}{cc}
  \boldsymbol{M_p}(\boldsymbol{q_p})   & \boldsymbol{0}  \\
  \boldsymbol{0}   &  \boldsymbol{M_r}(\boldsymbol{q_r})
\end{array}\right],
\boldsymbol{N}(\boldsymbol{q}, \dot{\boldsymbol{q}}) = \left[\begin{array}{c}
  \boldsymbol{N_p}(\boldsymbol{q}, \dot{\boldsymbol{q}})\\
  \boldsymbol{N_r}(\boldsymbol{q}, \dot{\boldsymbol{q}})
\end{array}\right]
$$

Linear discrete models should be obtained prior to developing LMPC controllers. In practice, when the robot is stable, the swing angles of the pendulum are usually minimal ($|\alpha| \leq 0.15 rad$ and $|\beta| \leq 0.25 rad$), so the model linearization can be conducted at the origin. Taking the transverse-axis model as an example:
\begin{equation}
    \dot{\boldsymbol{x}}_{\boldsymbol{r}} = \boldsymbol{A} \boldsymbol{x_r} + \boldsymbol{B} u_r + \boldsymbol{C}
\end{equation}

\noindent where $\setlength{\arraycolsep}{2pt} \boldsymbol{x_r} \!=\! \left[\begin{array}{cccc}
\beta & \dot{\beta} & \phi & \dot{\phi} \end{array}\right]^T$,
$\setlength{\arraycolsep}{2pt}
\left[\begin{array}{cc}
f_1  &  f_2  
\end{array} \right]^T \!=\! \boldsymbol{M_r}^{-1}(\boldsymbol{E_2}\tau_2 - \boldsymbol{N_r})$
$$
\setlength{\arraycolsep}{2pt}
\boldsymbol{A} \!=\! \left[\begin{array}{cccc}
0 & 1 & 0 & 0 \\
\frac{\partial f_{1}}{\partial \beta} & \frac{\partial f_{1}}{\partial \dot{\beta}} & \frac{\partial f_{1}}{\partial \phi} & \frac{\partial f_{1}}{\partial \dot{\phi}} \\
0 & 0 & 0 & 1 \\
\frac{\partial f_{2}}{\partial \beta} & \frac{\partial f_{2}}{\partial \dot{\beta}} & \frac{\partial f_{2}}{\partial \phi} & \frac{\partial f_{2}}{\partial \dot{\phi}}
\end{array}\right],
\boldsymbol{B} \!=\! \left[\begin{array}{c}
0 \\
\frac{\partial f_1}{\partial \tau_2} \\
0 \\
\frac{\partial f_2}{\partial \tau_2}
\end{array} \right],
\boldsymbol{C} \!=\! \left[\begin{array}{c}
0 \\
\left.f_{1}\right|_{\boldsymbol{x_r}=0} \\
0 \\
\left.f_{2}\right|_{\boldsymbol{x_r}=0}
\end{array} \right]
$$

Forward euler method is used to get the discrete model:
\begin{equation}
    \label{descrete}
    \boldsymbol{x_r}(k+1) = \boldsymbol{A_d}\boldsymbol{x_r}(k)+\boldsymbol{B_d}u_r(k)+\boldsymbol{C_d}
\end{equation}

\noindent in which $\boldsymbol{A_{d}}=\boldsymbol{I}+\boldsymbol{A} T_{s}$,
$\boldsymbol{B_{d}}=\boldsymbol{B} T_{s}$, 
$\boldsymbol{C_{d}}=\boldsymbol{C} T_{s}$

\subsection{Velocity Controller: ESO-MPC}

The largest issue with the spherical robot's velocity control is the mismatch of the model, which includes modeling errors and estimating error of $F_{fx}$ (the friction force along x axis is difficult to calculate or measure). However, when the robot's velocity is stable, we can approximate that the errors due to the model mismatch are constant, allowing us to employ the offset-free LMPC summarized in \cite{offset-freeLMPC}.
\begin{figure}[h]
\vspace{-0.3cm}
\centering
\setlength{\abovecaptionskip}{-0.1cm}
\includegraphics[width=1.0\linewidth]{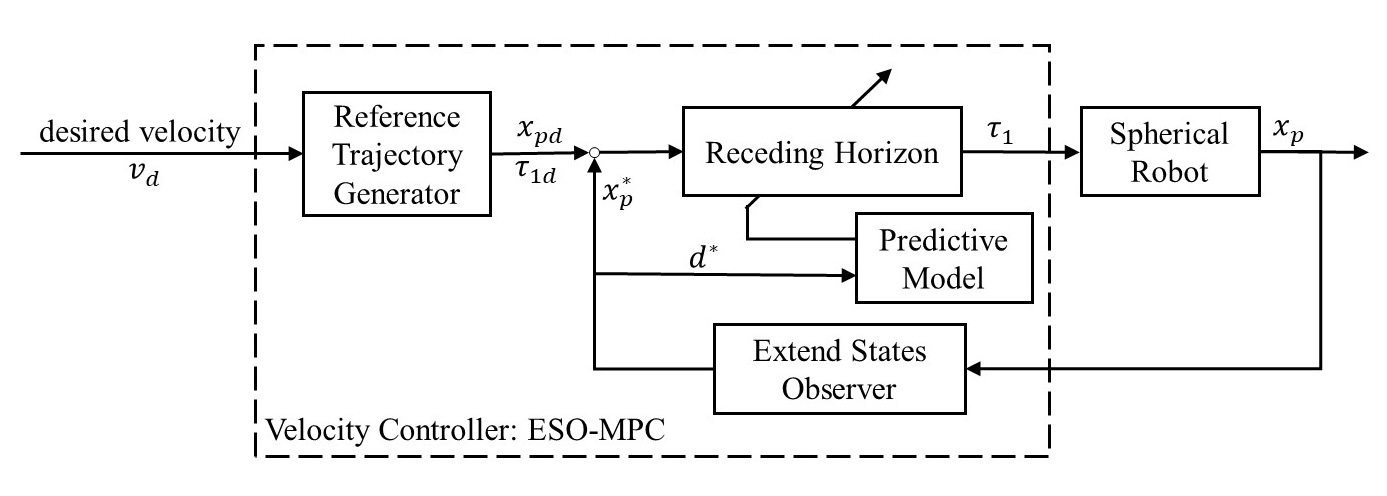}
\caption{The optimal velocity controller (ESO-MPC) for spherical robots.}
\label{Fig-ESOMPC}
\end{figure}

The main idea of ESO-MPC is to treat the errors caused by model mismatch as deterministic disturbances $\boldsymbol{d}$, then treat $\boldsymbol{d}$ as augmented states, design an extend states observer to get the optimal estimate of disturbances $\boldsymbol{d^*}$, then update the predictive model according to $\boldsymbol{d^*}$, and finally achieve unbiased control using LMPC. The structure of ESO-MPC is shown in Fig. \ref{Fig-ESOMPC} and the details can be found in our previous conference paper\cite{ESOMPC}.

\subsection{Reference Trajectory Generation in PWMPC}

\begin{figure}[h]
\vspace{-0.1cm}
\centering
\setlength{\abovecaptionskip}{-0.1cm}
\includegraphics[width=1.0\linewidth]{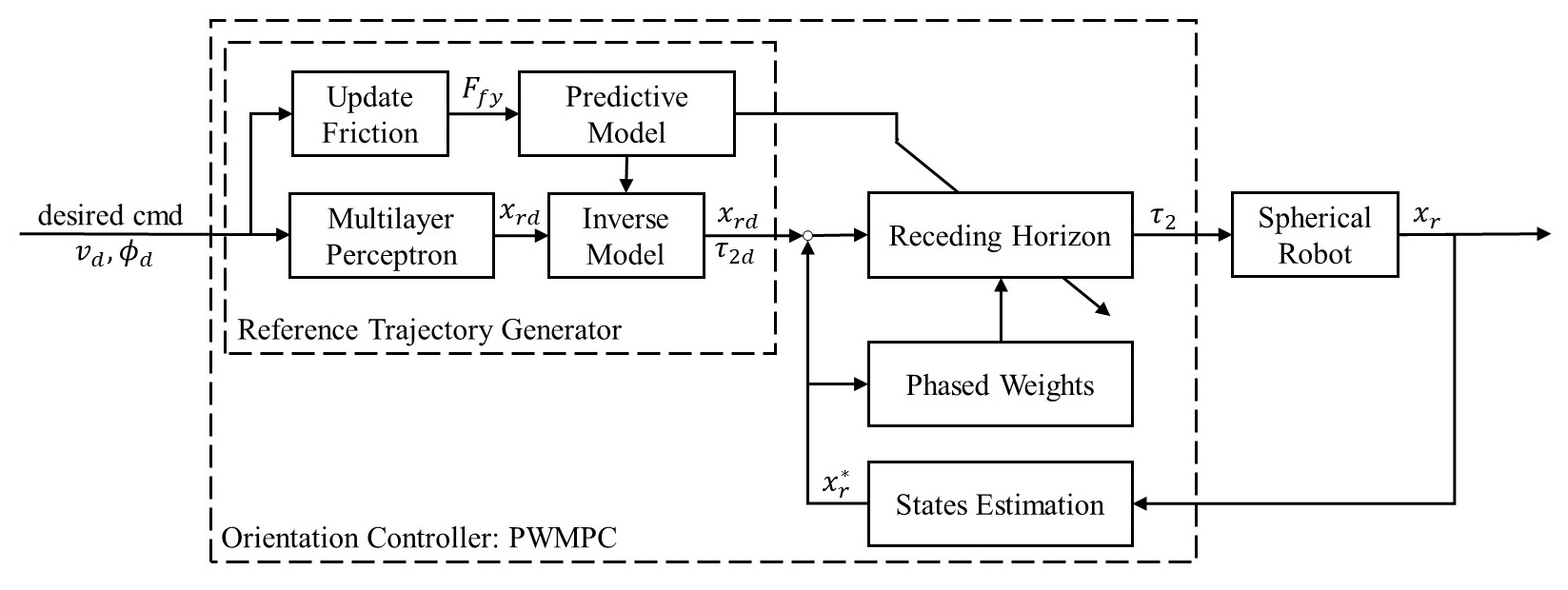}
\caption{The optimal orientation controller (PWMPC) for spherical robots.}
\label{Fig-PWMPC}
\vspace{-0.1cm}
\end{figure}

The structure of the MPC-based orientation controller for spherical robots is shown in Fig. \ref{Fig-PWMPC}. It can be seen that it mainly consists of an MLP-based reference trajectory generator and an MPC controller with time-varying weights, which is quite different from ESO-MPC.

Features of orientation control are different. In orientation control, the good news is that the friction along y axis can be calculated from centripetal force, as shown in \eqref{Ffy}, so we can update the friction to make the predictive model more accurate. However, there are two main challenges. The first is that the errors between predictive model and real system are not constant at stabilization, so the method used in velocity control cannot be applied for orientation control. The inability to update the model by estimating the errors means that the modeling errors cannot be eliminated, putting a higher demand on the accuracy of the reference trajectory.
\begin{equation}
    \label{Ffy}
    F_{fy} = \frac{(m_s+m_f+m_p)*v^2}{R}
\end{equation}

\noindent where $R$ is the turning radius, and $R = r/\tan{\phi}$

The reference trajectory consists of two parts: desired states $\boldsymbol{x_{rd}}$ and desired input $u_{rd}$, in which $\dot{\beta}_d$ and $\dot{\phi}_d$ can be set to zero, $\phi_d$ is obtained from the planning module, and $u_{rd}$ can be obtained by the inverse model after getting all the target states, so the key to generate the reference trajectory lies in $\beta_d$. The Newton-Euler method analysis reveals that $\beta_d$ is a function of both $v_d$ and $\phi_d$. To acquire the most accurate $\beta_d$ as possible, we collect 90 sets of data from the robot at different $v$ and $\phi$ and adopt MLP to get $\beta_d$. MLP is the most common neural network, and it has shown in \cite{MLP} that MLP can be trained to approximate virtually any smooth, measurable function.

\begin{figure}[h]
\centering
\setlength{\abovecaptionskip}{-0.3cm}
\includegraphics[width=0.8\linewidth]{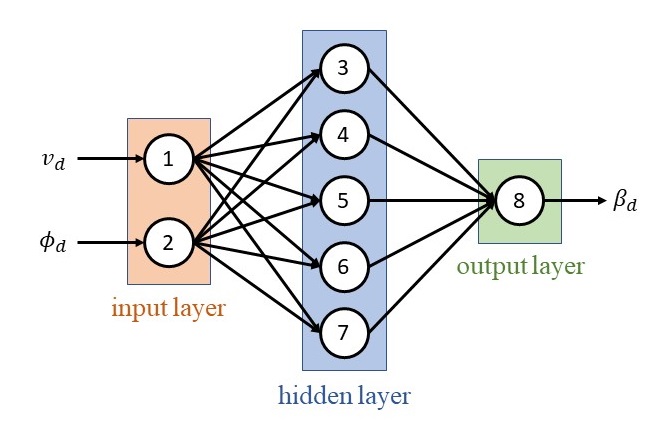}
\caption{The MLP used in this work.}
\label{Fig-MLP}
\end{figure}

The MLP used in this work is shown in Fig. \ref{Fig-MLP}. The Levenberg-Marquardt algorithm is used for training. At the same time, to avoid overfitting, the data is randomly divided into three parts: the training set (75\%), validation set (15\%), and test set (15\%). Finally, the transfer functions selected for the layers are the hyperbolic tangent (tansig) for hidden layer and the linear function for output layer. The best validation performance is shown at epoch 209, and the mean square error is $1.30 \times 10^{-8}$.
\vspace{-0.3cm}
\vspace{-0.3cm}

\subsection{Phased Change of Weights in PWMPC}

\begin{figure}[h]
\vspace{-0.8cm}
\centering
\setlength{\abovecaptionskip}{-0.1cm}
\includegraphics[width=0.8\linewidth]{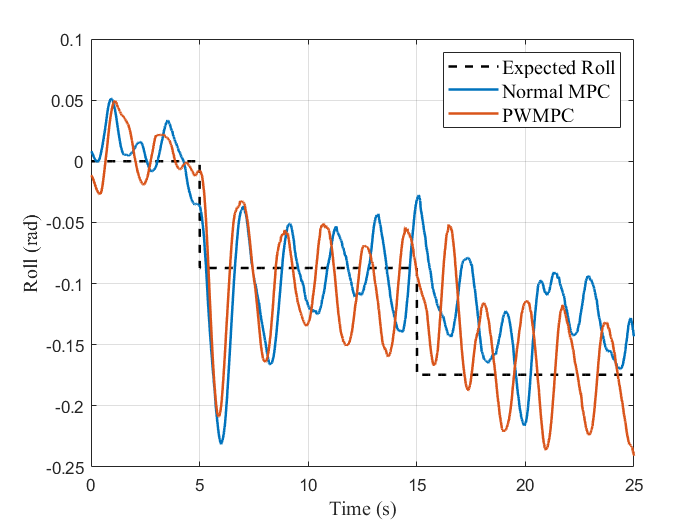}
\caption{Comparison of normal MPC and PWMPC ( the robot's velocity is set to be 1m/s during the experiments)}
\label{Fig-compare}
\vspace{-0.5cm}
\end{figure}

The second challenge of orientation control is that due to the time lag in the control of spherical robots, it is hard to find the balance between overshoot and settling time by adjusting weights when using normal MPC. As illustrated in Fig. \ref{Fig-compare}, when the target is received for the first time, the roll under the control of normal MPC has a considerable overshoot, however, when the target is changed, the normal MPC does not make the roll converge even after 10 seconds.

As a result, in the orientation control of spherical robots, we need to build the controller more thoroughly based on the robot's features. Inspired from the application of MPC in autonomous vehicles \cite{phasedweight}, we propose PWMPC, which divides the response process into three phases and assigns different weights to each phase: (1) Fast response phase. At the beginning of receiving a new target, the angles (e.g. $\beta$, $\phi$) are weighted more, so the controller responses rapidly. (2) Reduce overshoot phase. To reduce overshoot, the weights of the angular velocities (e.g. $\dot{\beta}$, $\dot{\phi}$) are increased as the angles reach the targets. (3) Stabilization phase. When approaching the steady state, the weights of each states are readjusted to maintain the robot's orientation. The performance of PWMPC is also shown in Fig. \ref{Fig-compare}, and it can be seen that not only the overshoot is slightly reduced, but also the convergence of $\phi$ is much faster when the target changes.

\subsection{QP Formulation of PWMPC}

In normal MPC, an optimization problem of the following form is solved on each iteration to get the optimal control sequence at the current time:
\begin{equation}
    \label{normalMPC}
    \begin{split}
    \min _{\boldsymbol{x}, \boldsymbol{u}} \sum_{i=1}^{N_p-1}&\left\|\boldsymbol{x}_{i}-\boldsymbol{x}_{i,ref}\right\|_{\boldsymbol{Q}}^2 + \sum_{i=1}^{N_c}\left\|\boldsymbol{u}_{i}-\boldsymbol{u}_{i,ref}\right\|_{\boldsymbol{R}}^2 \\
    + &\left\| \boldsymbol{x}_{N_p}-\boldsymbol{x}_{N_p,ref}\right\|_{\boldsymbol{P}}^2         
    \end{split}
\end{equation}
\vspace{-0.5cm}
\begin{align}
\text{s.t.} \qquad \qquad \qquad \boldsymbol{x}_{k+1} &= f(\boldsymbol{x}_{k}, \boldsymbol{u}_{k}) \tag{\ref{normalMPC}{a}} \\
\boldsymbol{x}_{i} \in \mathbb{X}, \quad &\boldsymbol{x}_{N_p} \in \mathbb{X}_f, \quad \boldsymbol{u}_{i} \in \mathbb{U} \tag{\ref{normalMPC}{b}}
\end{align}

\noindent where $\| \cdot \|$ is the Euclidean norm,  $f(\cdot)$ is the predictive model, $\boldsymbol{Q}$, $\boldsymbol{R}$ and $\boldsymbol{P}$ are weighting matrices, $N_p$ is the prediction horizon and $N_u$ is the control horizon. $\left\| \boldsymbol{x}_{N_p}-\boldsymbol{x}_{N_p,ref}\right\|_{\boldsymbol{P}}^2$ and $\mathbb{X}_f$ are the terminal cost and the terminal set, respectively, designed for the algorithm's stability \cite{MPCstable}.

Since the predictive model of PWMPC is a linear model shown in \eqref{descrete}, we can calculate the states sequence $\boldsymbol{X}$ from the initial states $\boldsymbol{x}_0$ and the control sequence $\boldsymbol{U}$:
\begin{equation}
    \label{getX}
    \boldsymbol{X} = \boldsymbol{A_{qp}} \boldsymbol{x}_0 + \boldsymbol{B_{qp}} \boldsymbol{U} + \boldsymbol{C_{qp}}
\end{equation}

Therefore, in PWMPC, the optimization problem defined in \eqref{normalMPC} can be transformed into a quadratic programming (QP) problem related only to $\boldsymbol{U}$:
\begin{equation}
    \label{QP}
    \min_{\boldsymbol{U}} \frac{1}{2}\boldsymbol{U}^T\boldsymbol{H}\boldsymbol{U}+\boldsymbol{U}^T\boldsymbol{f} 
\vspace{-0.55cm}
\end{equation}
\begin{align}
    \text{s.t.} \qquad \underline{\boldsymbol{d}} \leq \boldsymbol{D}\boldsymbol{U} \leq \overline{\boldsymbol{d}} \tag{\ref{QP}{a}}
\end{align}

\noindent in which $\boldsymbol{H}$, $\boldsymbol{f}$ and $\boldsymbol{D}$, respectively, are the Hessian matrix, the Jacobi matrix and the constraint matrix, and
$$
\boldsymbol{H} = 2(\boldsymbol{B_{qp}}^T\boldsymbol{Q_{qp}}\boldsymbol{B_{qp}} + \boldsymbol{R_{qp}})
\vspace{-0.1cm}
$$
$$
\boldsymbol{f} = 2[\boldsymbol{B_{qp}}^T\boldsymbol{Q_{qp}}(\boldsymbol{A_{qp}}\boldsymbol{x}_0+\boldsymbol{C_{qp}}-\boldsymbol{X}_{ref}) - \boldsymbol{R_{qp}}\boldsymbol{U}_{ref}]
$$

When implementing the algorithms, the arising QP problems are solved by the open-source solver qpOASES \cite{qpoases}. The average solution time for QP problems is tested to be $7.61\times10^{-5}$s when $N_p = 100$ and $N_c = 20$, and even when the time for preprocessing is added, the average computation time is $3.36\times10^{-3}$s, fully satisfying the requirement of control frequency.

\section{EXPERIMENTAL RESULTS}

To  verify the performance of the motion control framework proposed in this paper, a series of physical experiments were conducted. We first validated the effectiveness of individual controllers working independently. In subsection A, we fixed the robot's roll angle (let $\phi = 0$) and controlled the robot to track different velocities using ESO-MPC controller, with which we compared the traditional PID controller and our previously proposed HSMC controller \cite{hsmc1}. In subsection B, the robot's velocity was fixed, and we conducted a number of orientation control experiments using controllers such as Fuzzy-PID \cite{fuzzy}, HTSMC \cite{hsmc3}, and PWMPC. Then in subsection C, we tested the control frameworks (two controllers working simultaneously) including PID-based framework, SMC-based framework, and MPC-based framework proposed in this study.

As shown in Fig. \ref{Fig-robot}, an autonomous spherical robot was built for the experiments. The velocity and attitude of the robot are obtained through the encoders and IMUs. The algorithms are deployed on an Intel mini PC (Intel i7-8559 U, 2.70 GHz, Quad-core 64-bit) with ROS installed. The calculated torques are transmitted to a TI's digital signal processor (TMS320F28069) to control the motors and the control frequency is 50 Hz.  

\subsection{Velocity Control Experiments}

In order to verify the performance and robustness of the controllers for different velocities and terrains, we conducted ``constant velocity tracking'' experiments. Meanwhile, ``variable velocities tracking'' experiments and ``velocity tracking when turning'' experiments were carried out to compare the controllers' performance in actual use.
\begin{figure}[h]
\centering
\setlength{\abovecaptionskip}{-0.1cm}
\subfigure[curve of velocity]{
\includegraphics[width=0.48\linewidth]{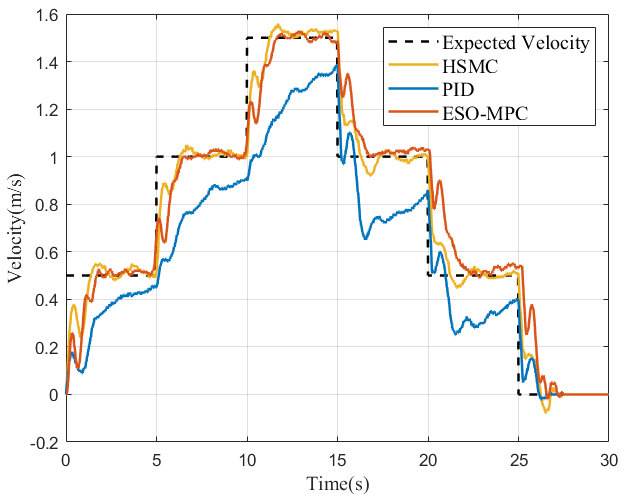}
}
\hspace{-5mm}
\subfigure[curve of pitch]{
\includegraphics[width=0.48\linewidth]{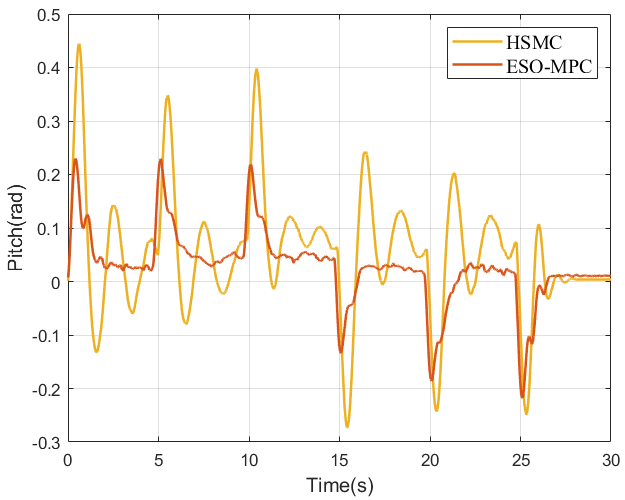}
}
\caption{Variable velocity tracking \cite{ESOMPC}}
\label{v}
\vspace{-0.1cm}
\end{figure}

The experimental results show that the rapidity and accuracy of HSMC and ESO-MPC controllers are significantly better than PID controller, and the robot's attitude as well as the motor's current are more stable under the control of ESO-MPC. As the specific experimental results and analysis described in our previous conference paper \cite{ESOMPC}, the ESO-MPC controller's settling time is 65\% shorter than that of PID, and the average change rate of the attitude and current when using ESO-MPC is 50\% of that with HSMC. Generally, ESO-MPC controller is a more comprehensive and effective velocity controller for spherical robots than PID and HSMC.

\subsection{Orientation Control Experiments}

The pendulum-driven spherical robots adjust its orientation by controlling the roll angle, therefore, the orientation control of spherical robots is essentially the roll angle ($\phi$) control. We performed the following roll angle control experiments to verify the performance of the orientation controllers. 

\subsubsection{Constant Roll Angle Tracking Experiments} 
\begin{figure}[h]
\centering
\setlength{\abovecaptionskip}{-0.1cm}
\subfigure[curve of roll]{
\includegraphics[width=0.48\linewidth]{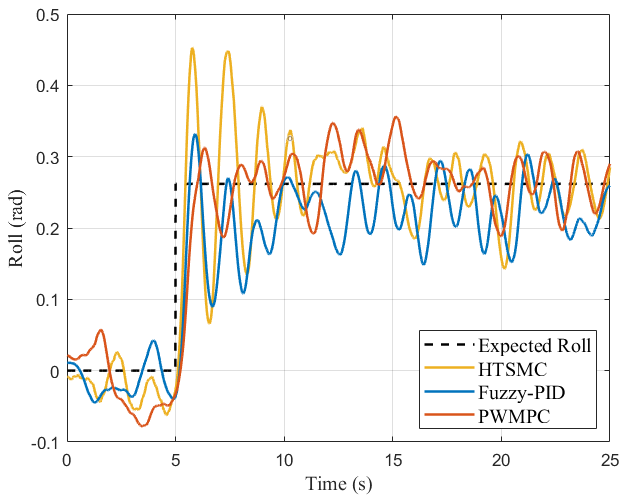}
}
\hspace{-5mm}
\subfigure[curve of current]{
\includegraphics[width=0.48\linewidth]{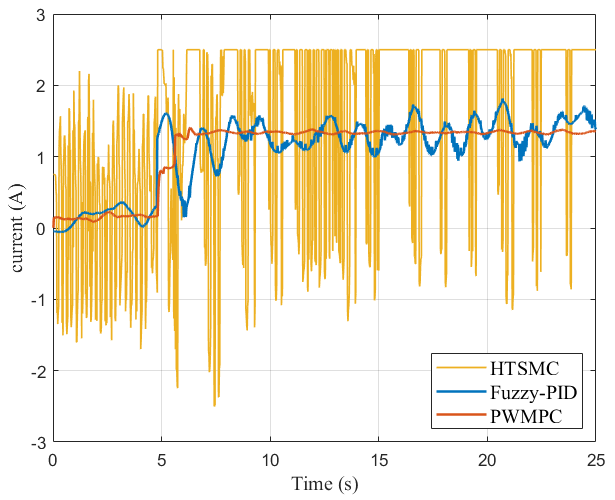}
}
\caption{Constant roll angle tracking ($v = 0.5 m/s$ and $\phi_d = 0.2618 rad$)}
\label{Fig-05_15}
\vspace{-0.7cm}
\end{figure}

\begin{figure}[h]
\centering
\setlength{\abovecaptionskip}{-0.1cm}
\subfigure[curve of roll]{
\includegraphics[width=0.48\linewidth]{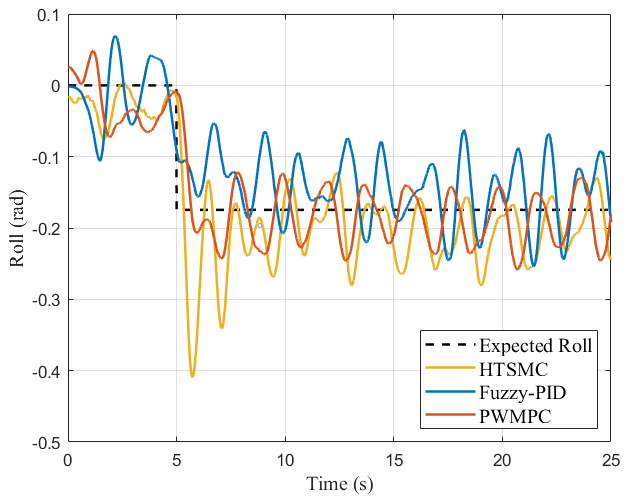}
}
\hspace{-5mm}
\subfigure[curve of current]{
\includegraphics[width=0.48\linewidth]{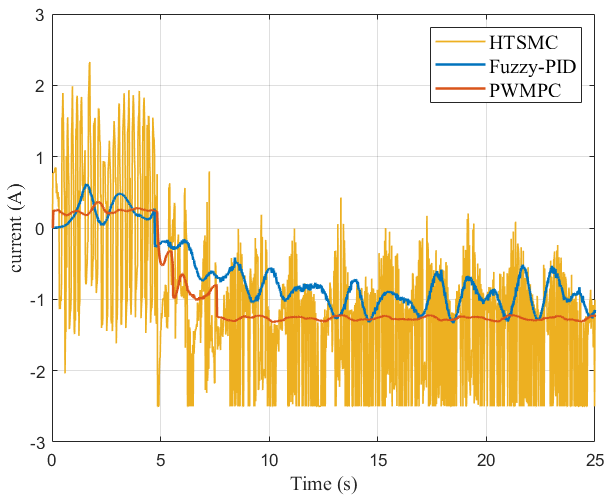}
}
\caption{Constant roll angle tracking ($v = 1.0 m/s$ and $\phi_d = -0.1745 rad$)}
\label{Fig-10_-10}
\vspace{-0.2cm}
\end{figure}

In these experiments, we tested each controller's ability in controlling the robot reach the target roll angle at different velocities. To ensure consistency, we let the robot travel straight for 5s before turning in each experiment. Fig. \ref{Fig-05_15} and \ref{Fig-10_-10} depict some of the results, and the data are shown in Table \ref{Indicators1}.

To compare the performance of controllers more precisely, we selected 8 indicators, including rise time $t_r$, overshoot $\sigma$, settling time $t_s$, root mean square error of roll in steady-state process $e_{rmse}$, variation range of angular velocity of roll $\dot{\phi}$, average absolute value of angular velocity $\dot{\phi}_{aa}$, energy consumption $Q$, and average absolute value of current change $\dot{I}_{aa}$. The first four indicators show the controller's rapidity and accuracy, while the last four reflect the stability of the robot's attitude and current during control. 

\begin{table*}[t]
\setlength{\tabcolsep}{5pt}
\setlength{\abovecaptionskip}{-0.2cm}
\caption{results of constant roll angle tracking experiments}
\label{Indicators1}
\renewcommand\arraystretch{1.5}
\begin{center}
\begin{tabular}{ccccccccccc}
\hline
\multirow{2}{*}{\textbf{\begin{tabular}[c]{@{}c@{}}Velocity\\ (m/s)\end{tabular}}} & \multirow{2}{*}{\textbf{\begin{tabular}[c]{@{}c@{}}Target Roll\\ (rad)\end{tabular}}} & \multirow{2}{*}{\textbf{Controller}} & \multicolumn{8}{c}{\textbf{Indicators}}                                                                                                                               \\ \cline{4-11} 
                                                                                   &                                                                                       &                                      & $\boldsymbol{t_r(s)}$ & $\boldsymbol{\sigma(\%)}$ & $\boldsymbol{t_s(s)}$ & $\boldsymbol{e_{rmse}(rad)}$ & $\boldsymbol{\dot{\phi}(rad/s)}$ & $\boldsymbol{\dot{\phi}_{aa}(rad/s)}$ & $\boldsymbol{Q(J)}$ & $\boldsymbol{\dot{I}_{aa}(A/s)}$ \\ \hline
\multirow{6}{*}{0.5}                                                               & \multirow{3}{*}{0.1745}                                                               & Fuzzy-PID                            & 0.64           & 5.86               & 7.89           & 0.0370                & {[}-0.2702, 0.3817{]}            & \textbf{0.0896}            & 828.74        & 1.25            \\
                                                                                   &                                                                                       & HTSMC                                & \textbf{0.32}           & 66.40              & 2.14           & \textbf{0.0258}                & {[}-0.6059, 0.7414{]}            & 0.3380            & 1660.04       & 50.53           \\
                                                                                   &                                                                                       & PWMPC                                & 0.58           & \textbf{0.00}                  & \textbf{0.66}           & 0.0269                & \textbf{{[}-0.2342, 0.3842{]}}            & 0.2093            & \textbf{775.94}        & \textbf{0.09}            \\ \cline{2-11} 
                                                                                   & \multirow{3}{*}{0.2618}                                                               & Fuzzy-PID                            & 0.60           & 26.64              & 9.39           & 0.0522                & {[}-0.4958, 0.6876{]}            & \textbf{0.1744}            & \textbf{1238.15}       & 1.89            \\
                                                                                   &                                                                                       & HTSMC                                & \textbf{0.43}           & 72.93              & 4.19           & 0.0400                & {[}-0.8695, 1.0413{]}            & 0.4396            & 1992.75       & 20.05           \\
                                                                                   &                                                                                       & PWMPC                                & 0.86           & \textbf{0.00}                  & \textbf{1.00}           & \textbf{0.0383}                & \textbf{{[}-0.2654, 0.5123{]}}            & 0.2974            & 1270.95       & \textbf{0.14}            \\ \hline
\multirow{6}{*}{1.0}                                                                 & \multirow{3}{*}{-0.0873}                                                              & Fuzzy-PID                            & 0.22           & 170.38             & 12.48          & \textbf{0.0249}                & {[}-0.5770, 0.6648{]}            & 0.2813            & \textbf{372.25}        & 1.38            \\
                                                                                   &                                                                                       & HTSMC                                & 0.23           & 167.79             & 4.49           & 0.0259                & {[}-0.5848, 0.5694{]}            & 0.2469            & 798.20        & 43.26           \\
                                                                                   &                                                                                       & PWMPC                                & \textbf{0.15}           & \textbf{100.12}             & \textbf{1.60}           & 0.0304                & \textbf{{[}-0.3486, 0.2391{]}}            & \textbf{0.1234}            & 593.89        & \textbf{0.47}            \\ \cline{2-11} 
                                                                                   & \multirow{3}{*}{-0.1745}                                                              & Fuzzy-PID                            & 3.02           & 18.76              & 7.16           & 0.0532                & {[}-0.4744, 0.4814{]}            & \textbf{0.1153}            & \textbf{827.38}        & 1.49            \\
                                                                                   &                                                                                       & HTSMC                                & \textbf{0.31}           & 134.70             & 2.34           & 0.0469                & {[}-0.9320. 0.6188{]}            & 0.4060            & 1362.96       & 45.48           \\
                                                                                   &                                                                                       & PWMPC                                & 0.63           & \textbf{0.00}                  & \textbf{0.68}           & \textbf{0.0386}                & \textbf{{[}-0.4193, 0.3119{]}}            & 0.2432            & 1156.88       & \textbf{0.62}            \\ \hline
\end{tabular}
\end{center}
\vspace{-0.3cm}
\end{table*} 

It is obvious from the figures that the rapidity of HTSMC and PWMPC is significantly better than that of Fuzzy-PID, and PWMPC has a great benefit over HTSMC in terms of overshoot and current stability. According to Table \ref{Indicators1}, the settling time $t_s$ of PWMPC is greatly reduced (by 70\% compared to HTSMC), and the stability of the robot's attitude during turning is also significantly improved under the control of PWMPC (the variation range of $\dot{\phi}$ and the value of $\dot{\phi}_{aa}$ is reduced by 50\% compared to HTSMC).

\subsubsection{Variable Roll Angle Tracking Experiments} To test the effectiveness of each controller in actual use, we designed the following experiments, including multiple-steps tracking experiments (the target roll changes every 5 seconds) and sine-wave tracking experiments (the function of target roll is shown in \eqref{sin-wave}). The experimental results of multiple-steps tracking and sine-wave tracking are displayed in Fig. \ref{Fig-step} and \ref{Fig-fixv_sinroll}, and more details are shown in Table \ref{Indicators2}.
\begin{equation}
    \label{sin-wave}
    \phi_{d}=\left\{\begin{array}{cc}
0 & , t<2.0s \\
10 \sin (0.15 t-0.3) & , t\geq 2.0s
\end{array}\right.
\end{equation}

From the figures, it can be seen that PWMPC can handle the problem of variable roll tracking well. In addition, PWMPC also gets a more stable current without large fluctuations while tracking. Combined with the data in Table \ref{Indicators2}, we can find that under the control of PWMPC, $e_{rmse}$ is reduced by 30\% relative to Fuzzy-PID for both discrete and continuous changes of the target, and $\dot{\phi}_{aa}$ is also reduced compared to HTSMC (by 40\% in multiple-steps tracking experiments and 20\% in sine-wave tracking experiments).

Based on the results, it is clear that PWMPC is a more suitable orientation controller for spherical robots, which can track the target quickly and precisely with less impact on the robot's attitude, and the output current is more stable.
\begin{table}[h]
\vspace{-0.1cm}
\setlength{\abovecaptionskip}{-0.35cm}
\caption{results of variable roll angle tracking experiments}
\label{Indicators2}
\renewcommand\arraystretch{1.5}
\begin{center}
\setlength{\tabcolsep}{0.5mm}{
\begin{tabular}{cccccc}
\hline
\multirow{3}{*}{\textbf{Experiment}} & \multirow{3}{*}{\textbf{Controller}} & \multicolumn{4}{c}{\textbf{Indicators}}                                      \\ \cline{3-6} 
                                     &                                      & \multirow{2}{*}{\begin{tabular}{c}
                                        $\boldsymbol{e_{rmse}}$ \\
                                        $\boldsymbol{(rad)}$ 
                                     \end{tabular}} &
                                     \multirow{2}{*}{\begin{tabular}{c}
                                        $\boldsymbol{\dot{\phi}}$ \\
                                        $\boldsymbol{(rad/s)}$ 
                                     \end{tabular}}    &
                                     \multirow{2}{*}{\begin{tabular}{c}
                                        $\boldsymbol{\dot{\phi}_{aa}}$  \\
                                        $\boldsymbol{(rad/s)}$
                                     \end{tabular}} & 
                                     \multirow{2}{*}{\begin{tabular}{c}
                                        $\boldsymbol{\dot{I}_{aa}}$  \\
                                        $\boldsymbol{(A/s)}$
                                     \end{tabular}} \\ 
                                     &  &   &   &   & \\\hline
\multirow{3}{*}{\begin{tabular}[c]{@{}c@{}}Multiple-steps\\ Tracking\\ \end{tabular}}                    & Fuzzy-PID                            & 0.0740         & {[}-0.5730, 0.5010{]} & 0.1638           & 1.28             \\
                                     & HTSMC                                & 0.0676         & {[}-1.2982, 1.1287{]} & 0.2242           & 37.02            \\
                                     & PWMPC                                & \textbf{0.0592}         & \textbf{{[}-0.4190, 0.5444{]}} & \textbf{0.1303}           & \textbf{0.67}             \\ \hline
\multirow{3}{*}{\begin{tabular}[c]{@{}c@{}}Sine-wave\\ Tracking\\ \end{tabular}}                    & Fuzzy-PID                            & 0.0576         & {[}-0.5506, 0.5143{]} & 0.1236           & 1.14             \\
                                     & HTSMC                                & \textbf{0.0383}         & {[}-0.4944, 0.4842{]} & 0.1193           & 44.61            \\
                                     & PWMPC                                & 0.0388         & \textbf{{[}-0.3084, 0.4130{]}} & \textbf{0.0992}           & \textbf{0.52}             \\ \hline
\end{tabular}
}
\end{center}
\vspace{-0.5cm}
\end{table}
\begin{figure}[h]
\vspace{-0.2cm}
\centering
\setlength{\abovecaptionskip}{-0.1cm}
\subfigure[curve of roll]{
\includegraphics[width=0.48\linewidth]{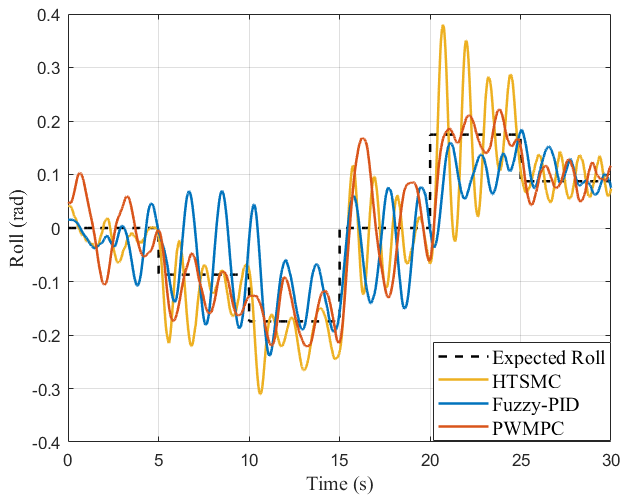}
}
\hspace{-5mm}
\subfigure[curve of current]{
\includegraphics[width=0.48\linewidth]{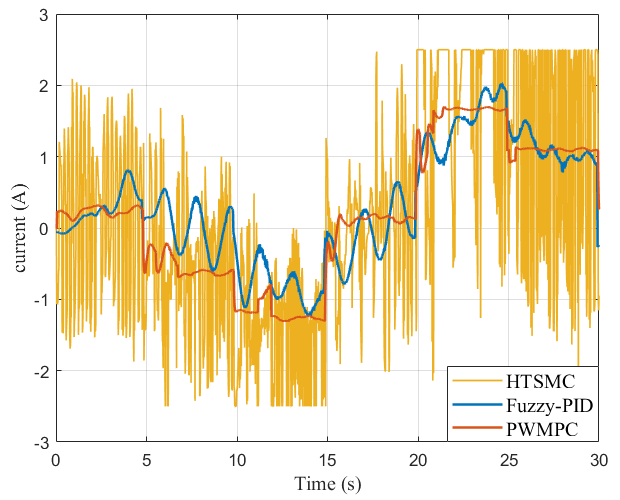}
}
\caption{Multiple-steps tracking}
\label{Fig-step}
\vspace{-0.3cm}
\end{figure}
\begin{figure}[h]
\centering
\setlength{\abovecaptionskip}{-0.1cm}
\subfigure[curve of roll]{
\includegraphics[width=0.48\linewidth]{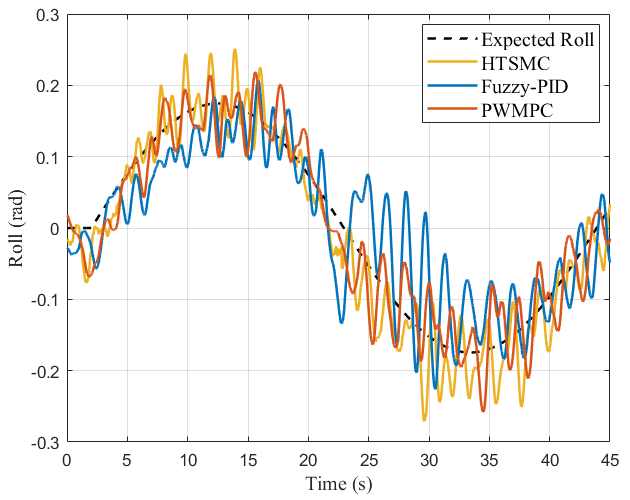}
}
\hspace{-5mm}
\subfigure[curve of current]{
\includegraphics[width=0.48\linewidth]{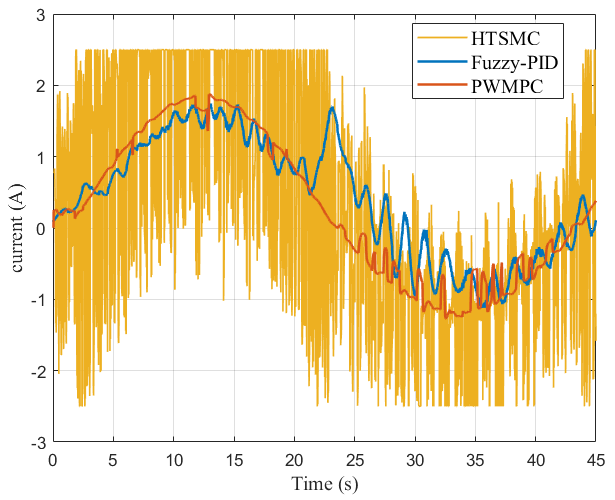}
}
\caption{Sine-wave tracking}
\label{Fig-fixv_sinroll}
\vspace{-0.4cm}
\end{figure}

\begin{figure*}[t]
\centering
\setlength{\abovecaptionskip}{-0.1cm}
\subfigure[curve of velocity]{
\includegraphics[width=0.32\linewidth]{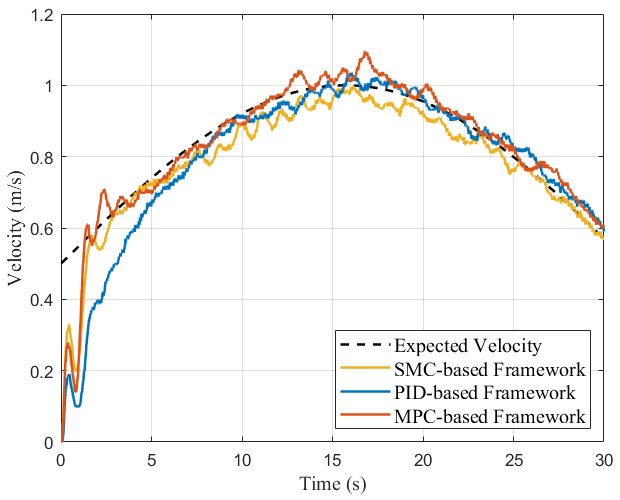}
}
\hspace{-5mm}
\subfigure[curve of roll]{
\includegraphics[width=0.32\linewidth]{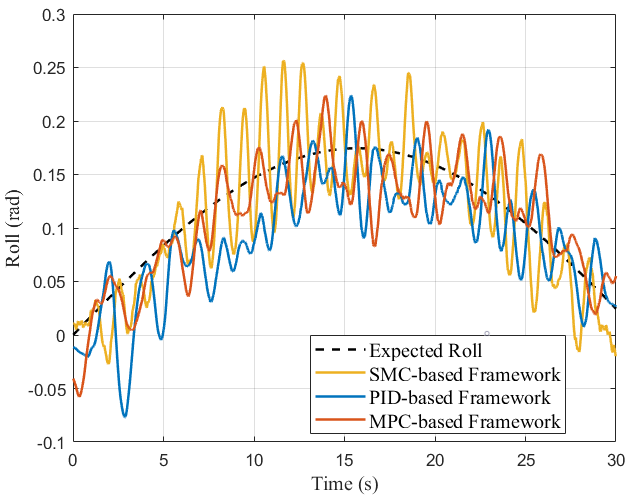}
}
\hspace{-5mm}
\subfigure[curve of transverse-axis motor's current]{
\includegraphics[width=0.32\linewidth]{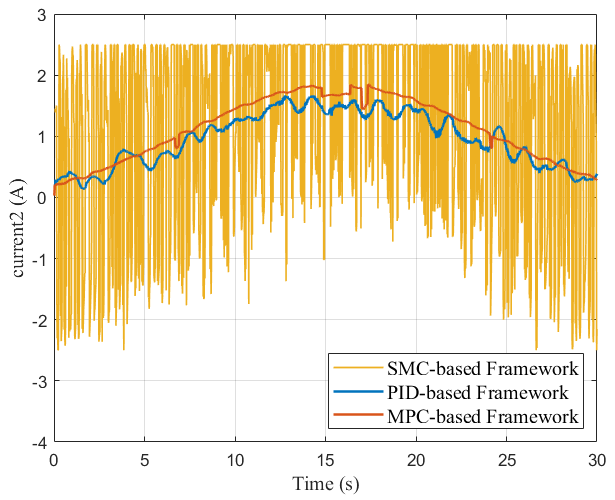}
}
\caption{Simultaneous control experiments}
\label{Fig-sinv_sinroll}
\vspace{-0.3cm}
\end{figure*}

\subsection{Motion Control Experiments}

Although we have verified the superiority of ESO-MPC and PWMPC in velocity control and orientation control respectively through previous experiments, we need to further validate the performance of the whole control framework because the transverse and longitudinal control of spherical robots are coupled. In this section, we designed simultaneous control experiments as well as trajectory tracking experiments to test the control frameworks. The control frameworks for comparison are PID-based (velocity controller: PID, orientation controller: Fuzzy-PID), SMC-based (velocity controller: HSMC, orientation controller: HTSMC), and MPC-based (velocity Controller: ESO-MPC, orientation controller: PWMPC).

\subsubsection{Simultaneous Control Experiments} In these experiments, both the target velocity and the target roll angle are time-varying, and both are sine-wave like functions, as shown in \eqref{sinv-sinroll}. The curves of some states are shown in Fig. \ref{Fig-sinv_sinroll}, and Table \ref{Indicators3} presents the specific results, where $\dot{\theta}$ indicates the variation range of angular velocity of pitch, $\dot{\theta}_{aa}$ is the average absolute value of angular velocity, $\dot{I}_{1aa}$ and $\dot{I}_{2aa}$ denote the average absolute values of longitudinal-axis and transverse-axis motor's current variation, respectively.
\begin{equation}
    \label{sinv-sinroll}
    \left\{\begin{array}{l}
v_d = 0.5\sin{(0.1t)}+0.5 \\
\phi_d = 0.1745\sin{(0.1t)}
\end{array}\right.
\end{equation}

As can be seen in Fig. \ref{Fig-sinv_sinroll}, both the SMC-based and MPC-based control frameworks respond faster and more accurately for both velocity and orientation control, and the MPC-based control framework still has a significant advantage in terms of current stability. When combined with the indicators in Table \ref{Indicators3}, the MPC-based framework is shown to be optimal in almost all aspects, including root mean square errors, attitude variation, and current variation.
\begin{table}[h]
\vspace{-0.1cm}
\setlength{\abovecaptionskip}{-0.01cm}
\caption{results of simultaneous control experiments}
\label{Indicators3}
\renewcommand\arraystretch{1.5}
\setlength{\tabcolsep}{0.3mm}{
\begin{tabular}{ccccc}
\hline
\multirow{2}{*}{\textbf{\begin{tabular}[c]{@{}c@{}}Control\\ Framework\end{tabular}}} & \multicolumn{4}{c}{\textbf{Indicators (related to velocity control)}}                         \\ \cline{2-5} 
                                                                                      & $\boldsymbol{e_{rmse}(m/s)}$ & $\boldsymbol{\dot{\theta}(rad/s)}$ & $\boldsymbol{\dot{\theta}_{aa}(rad/s)}$ & $\boldsymbol{\dot{I}_{1aa}(A/s)}$ \\ \hline
PID-based                                                                                   & 0.1071                & \textbf{{[}-0.3037, 0.5246{]}} & \textbf{0.0601}                    & 1.05             \\
SMC-based                                                                                   & 0.0741                & {[}-0.8122, 0.9659{]} & 0.0945                    & 1.93             \\
MPC-based                                                                                   & \textbf{0.0726}                & {[}-0.7113, 0.7470{]} & 0.0714                    & \textbf{1.00}             \\ \hline
\multirow{2}{*}{\textbf{\begin{tabular}[c]{@{}c@{}}Control\\ Framework\end{tabular}}} & \multicolumn{4}{c}{\textbf{Indicators (related to orientation control)}}                      \\ \cline{2-5} 
                                                                                      & $\boldsymbol{e_{rmse}(rad)}$ & $\boldsymbol{\dot{\phi}(rad/s)}$ & $\boldsymbol{\dot{\phi}_{aa}(rad/s)}$ & $\boldsymbol{\dot{I}_{2aa}(A/s)}$ \\ \hline
PID-based                                                                                   & 0.0420                & \textbf{{[}-0.2600, 0.2668{]}} & 0.0862                    & 0.94             \\
SMC-based                                                                                   & 0.0369                & {[}-0.5131, 0.5241{]} & 0.1536                    & 48.47            \\
MPC-based                                                                                   & \textbf{0.0310}                & {[}-0.3166, 0.2459{]} & \textbf{0.0777}                    & \textbf{0.27}             \\ \hline
\end{tabular}
}
\end{table}

\subsubsection{Trajectory Tracking Experiments} 
\begin{figure}[h]
\centering
\setlength{\abovecaptionskip}{-0.1cm}
\includegraphics[width=0.8\linewidth]{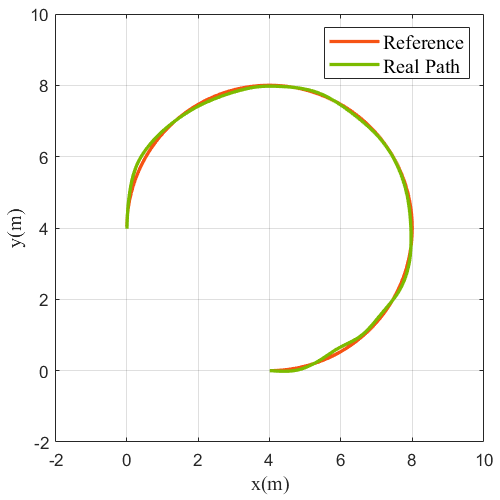}
\caption{Tracking a circular trajectory}
\label{Fig-circle}
\vspace{-0.3cm}
\end{figure}

To verify whether the MPC-based control framework can be adapted to the existing planning framework, we migrated it to the previously proposed trajectory tracking framework \cite{hsmc3} and replaced the controllers in that framework with the MPC-based controllers. Fig. \ref{Fig-circle} displays the outcomes of tracking a circular trajectory (the function of the reference trajectory is shown in \eqref{circle}) using the modified trajectory tracking framework.
\begin{equation}
    \label{circle}
    \left\{\begin{array}{ll}
x_{{ref}} & = 4\cos{(0.125t-0.5\pi)} + 4 \\
y_{{ref}} & = 4\sin{(0.125t-0.5\pi)} + 4 \\
\psi_{{ref}} & = 0.125t
\end{array}\right.
\end{equation}

The tracking results show that the modified trajectory tracking framework runs well, and the real path of the robot is close to the reference, demonstrating that the MPC-based motion control framework proposed in this study can be effectively applied to the existing system. 

\section{CONCLUSIONS}

In this paper, we discuss the motion control of pendulum-driven spherical robots in detail, and the effect of motion control on perception as well as on mechanical structure is considered for the first time, emphasizing the need to minimize the variation of robot's attitude and motor's current during control. Then an MPC-based optimal control framework for spherical robots is proposed, which uses MPC to systematically handle robot dynamics, constraints, and conflicting control objectives. 

To apply MPC, we decompose the whole-body model of the robot into a longitudinal-axis model and a transverse-axis model for computational efficiency, and design controllers for them respectively. Considering that the modelling errors are constant in velocity control, we combine ESO and MPC to design an optimal velocity controller ESO-MPC. In orientation control, we introduce MLP to generate reference trajectory, and use MPC with time-varying weights to achieve optimal control. So far, we have developed an MPC-based optimal motion control framework for spherical robots. 

The superiority of the MPC-based control framework is verified by a series of physical experiments. The experimental results illustrate that, whether it's controlling velocity or orientation, the MPC-based framework is much faster and more accurate than the PID-based framework, and it has significant advantages over the SMC-based framework in terms of overshoot, the stability of robot's attitude, and current stability. Additionally, the control framework designed in this study can be well adapted to the planning framework of the robot to achieve path tracking and trajectory tracking.

In this study, there are still some areas that can be further explored. For example, decomposing the whole model into several sub-models and then controlling the sub-models independently actually ignores the coupling of the sub-models. Although the final results are well, the optimal control of each sub-model does not always imply the optimal control of the whole system, so designing one control to realize the control of the whole system may be a better solution.






\bibliographystyle{IEEEtran}    
\bibliography{ref}

\end{document}